%% file: paper-all.tex
\documentclass{article}

\PassOptionsToPackage{numbers, compress}{natbib}
\usepackage[final]{nips_2016}

\usepackage[utf8]{inputenc} 
\usepackage[T1]{fontenc}    
\usepackage{hyperref}       
\usepackage{url}            
\usepackage{booktabs}       
\usepackage{amsfonts}       
\usepackage{nicefrac}       
\usepackage{microtype}      
\usepackage{multirow}

\usepackage[ruled,vlined]{algorithm2e}
\usepackage{amsmath, amsthm, amssymb, amsbsy, mathtools}
\usepackage{tabularx}

\usepackage{empheq}
\usepackage{tikz}
\usetikzlibrary{arrows,shapes,backgrounds,fit,positioning}
\usepackage{varwidth}

\newcommand{\bm}[1]{\boldsymbol{\mathbf{#1}}}
\input{bold_letters.tex}
\newcommand{\half}{\nicefrac{1}{2}}
\newcommand{\E}{\mathbb{E}}
\newcommand{\I}{\mathbf{I}}
\newcommand{\N}{\mathcal{N}}
\renewcommand{\O}{\mathcal{O}}
\newcommand{\R}{\mathbb{R}}
\newcommand{\0}{\mathbf{0}}
\DeclareMathOperator*{\argmin}{argmin}

\title{A scalable end-to-end Gaussian process adapter
for irregularly sampled time series classification}

\author{
  Steven Cheng-Xian Li
  \hfill
  Benjamin Marlin
  \\
  College of Information and Computer Sciences \\
  University of Massachusetts Amherst \\
  Amherst, MA 01003 \\
  \texttt{\{cxl,marlin\}@cs.umass.edu} \\
}

\begin{document}

\maketitle

\begin{abstract}
  \input{abstract.tex}

\end{abstract}

\input{introduction.tex}

\input{irregular-time-series.tex}
\input{gp-adapter.tex}
\input{fast_sampling.tex}
\input{end_to_end.tex}

\input{related.tex}
\input{experiments.tex}
\input{conclusion.tex}
\input{acknowledgements.tex}

{
\small
\bibliographystyle{plainnat}
\bibliography{paper}
}
\newpage
\clearpage
\appendix
\input{gradients.tex}
\input{cubic-interp.tex}

\input{arch.tex}

\end{document}

%% file: bold_letters.tex
\newcommand{\overridecommand}[2]{
  \providecommand{#1}{}
  \renewcommand{#1}{#2}
}

\overridecommand{\aa}{\mathbf{a}}
\overridecommand{\bb}{\mathbf{b}}
\overridecommand{\cc}{\mathbf{c}}
\overridecommand{\dd}{\mathbf{d}}
\overridecommand{\ee}{\mathbf{e}}
\overridecommand{\ff}{\mathbf{f}}
\overridecommand{\bmg}{\mathbf{g}}   
\overridecommand{\hh}{\mathbf{h}}
\overridecommand{\ii}{\mathbf{i}}
\overridecommand{\jj}{\mathbf{j}}
\overridecommand{\kk}{\mathbf{k}}
\overridecommand{\bml}{\mathbf{l}}   
\overridecommand{\mm}{\mathbf{m}}
\overridecommand{\nn}{\mathbf{n}}
\overridecommand{\oo}{\mathbf{o}}
\overridecommand{\pp}{\mathbf{p}}
\overridecommand{\qq}{\mathbf{q}}
\overridecommand{\rr}{\mathbf{r}}
\overridecommand{\ss}{\mathbf{s}}
\overridecommand{\tt}{\mathbf{t}}
\overridecommand{\uu}{\mathbf{u}}
\overridecommand{\vv}{\mathbf{v}}
\overridecommand{\ww}{\mathbf{w}}
\overridecommand{\xx}{\mathbf{x}}
\overridecommand{\yy}{\mathbf{y}}
\overridecommand{\zz}{\mathbf{z}}

\overridecommand{\BB}{\mathbf{B}}
\overridecommand{\CC}{\mathbf{C}}
\overridecommand{\DD}{\mathbf{D}}
\overridecommand{\EE}{\mathbf{E}}
\overridecommand{\FF}{\mathbf{F}}
\overridecommand{\GG}{\mathbf{G}}
\overridecommand{\HH}{\mathbf{H}}
\overridecommand{\II}{\mathbf{I}}
\overridecommand{\JJ}{\mathbf{J}}
\overridecommand{\KK}{\mathbf{K}}
\overridecommand{\LL}{\mathbf{L}}
\overridecommand{\MM}{\mathbf{M}}
\overridecommand{\NN}{\mathbf{N}}
\overridecommand{\OO}{\mathbf{O}}
\overridecommand{\PP}{\mathbf{P}}
\overridecommand{\QQ}{\mathbf{Q}}
\overridecommand{\RR}{\mathbf{R}}
\overridecommand{\SS}{\mathbf{S}}
\overridecommand{\TT}{\mathbf{T}}
\overridecommand{\UU}{\mathbf{U}}
\overridecommand{\VV}{\mathbf{V}}
\overridecommand{\WW}{\mathbf{W}}
\overridecommand{\XX}{\mathbf{X}}
\overridecommand{\YY}{\mathbf{Y}}
\overridecommand{\ZZ}{\mathbf{Z}}

\overridecommand{\aalpha}{\boldsymbol{\alpha}}
\overridecommand{\bbeta}{\boldsymbol{\beta}}
\overridecommand{\ggamma}{\boldsymbol{\gamma}}
\overridecommand{\ddelta}{\boldsymbol{\delta}}
\overridecommand{\eepsilon}{\boldsymbol{\epsilon}}
\overridecommand{\vvarepsilon}{\boldsymbol{\varepsilon}}
\overridecommand{\zzeta}{\boldsymbol{\zeta}}
\overridecommand{\eeta}{\boldsymbol{\eta}}
\overridecommand{\ttheta}{\boldsymbol{\theta}}
\overridecommand{\vvartheta}{\boldsymbol{\vartheta}}
\overridecommand{\iiota}{\boldsymbol{\iota}}
\overridecommand{\kkappa}{\boldsymbol{\kappa}}
\overridecommand{\llambda}{\boldsymbol{\lambda}}
\overridecommand{\mmu}{\boldsymbol{\mu}}
\overridecommand{\nnu}{\boldsymbol{\nu}}
\overridecommand{\xxi}{\boldsymbol{\xi}}
\overridecommand{\ppi}{\boldsymbol{\pi}}
\overridecommand{\vvarpi}{\boldsymbol{\varpi}}
\overridecommand{\rrho}{\boldsymbol{\rho}}
\overridecommand{\vvarrho}{\boldsymbol{\varrho}}
\overridecommand{\ssigma}{\boldsymbol{\sigma}}
\overridecommand{\vvarsigma}{\boldsymbol{\varsigma}}
\overridecommand{\ttau}{\boldsymbol{\tau}}
\overridecommand{\uupsilon}{\boldsymbol{\upsilon}}
\overridecommand{\pphi}{\boldsymbol{\phi}}
\overridecommand{\vvarphi}{\boldsymbol{\varphi}}
\overridecommand{\cchi}{\boldsymbol{\chi}}
\overridecommand{\ppsi}{\boldsymbol{\psi}}
\overridecommand{\oomega}{\boldsymbol{\omega}}
\overridecommand{\GGamma}{\boldsymbol{\Gamma}}
\overridecommand{\DDelta}{\boldsymbol{\Delta}}
\overridecommand{\TTheta}{\boldsymbol{\Theta}}
\overridecommand{\LLambda}{\boldsymbol{\Lambda}}
\overridecommand{\XXi}{\boldsymbol{\Xi}}
\overridecommand{\PPi}{\boldsymbol{\Pi}}
\overridecommand{\SSigma}{\boldsymbol{\Sigma}}
\overridecommand{\UUpsilon}{\boldsymbol{\Upsilon}}
\overridecommand{\PPhi}{\boldsymbol{\Phi}}
\overridecommand{\PPsi}{\boldsymbol{\Psi}}
\overridecommand{\OOmega}{\boldsymbol{\Omega}}

%% file: abstract.tex
We present a general framework for
classification of sparse and irregularly-sampled time series.
The properties of such time series can result in substantial uncertainty
about the values of the underlying temporal processes,
while making the data difficult to deal with using
standard classification methods that assume fixed-dimensional feature spaces.
To address these challenges, we propose an uncertainty-aware classification
framework based on a special computational
layer we refer to as the Gaussian process adapter that can connect
irregularly sampled time series data to any black-box classifier
learnable using gradient descent.
We show how to scale up the required computations
based on combining the structured kernel interpolation framework and
the Lanczos approximation method, and how to
discriminatively train the Gaussian process adapter in combination with a
number of classifiers end-to-end using backpropagation.

%% file: introduction.tex
\section{Introduction}
\label{sec:intro}

In this paper, we propose a general framework
for classification of sparse and irregularly-sampled time series.
An irregularly-sampled time series is a sequence of samples
with irregular intervals between their observation times.
These intervals can be large when the time series are also sparsely sampled. Such time series data are studied in various areas including
climate science \citep{schulz1997spectrum},
ecology \citep{clark2004population},
biology \citep{ruf1999lomb},
medicine \citep{marlin-ihi2012}
and astronomy \citep{scargle-astro1982}.
Classification in this setting is challenging both because the data cases are
not naturally defined in a fixed-dimensional feature space due to irregular
sampling and variable numbers of samples, and because there can be substantial
uncertainty about the underlying temporal processes due to the sparsity of
observations.

Recently, \citet{li2015classification} introduced
the mixture of expected Gaussian kernels (MEG) framework,
an uncertainty-aware kernel
for classifying sparse and irregularly sampled time series.
Classification with MEG kernels is shown to outperform
models that ignore uncertainty due to sparse and irregular sampling.
On the other hand, various deep learning models including convolutional neural networks \citep{lecun2004learning} have been successfully
applied to fields such as computer vision and natural language processing,
and have been shown to achieve state-of-the-art results on various tasks.
Some of these models have desirable properties for
time series classification, but cannot be directly applied to
sparse and irregularly sampled time series.

Inspired by the MEG kernel,
we propose an uncertainty-aware classification framework
that enables learning black-box classification models from sparse and
irregularly sampled time series data. This framework is based on the
use of a computational layer that we refer to as the Gaussian process (GP)
adapter. The GP adapter uses Gaussian process regression to transform the
irregular time series data into a uniform representation, allowing
sparse and irregularly sampled data to be fed into any black-box classifier
learnable using gradient descent while preserving uncertainty.
However, the $\O(n^3)$ time and $\O(n^2)$ space
of exact GP regression makes the GP adapter prohibitively expensive when scaling up to large time series.

To address this problem, we show how to speed up the key computation of
sampling from a GP posterior based on combining
the structured kernel interpolation (SKI)
framework that was recently proposed by \citet{wilson2015kernel}
with Lanczos methods for approximating
matrix functions \citep{chow2014preconditioned}.
Using the proposed sampling algorithm, the GP adapter can run in linear time and
space in terms of the length of the time series, and
$\O(m\log m)$ time when $m$ inducing points are used.

We also show that GP adapter can be trained end-to-end together with
the parameters of the chosen classifier by backpropagation
through the iterative Lanczos method. We present results using logistic
regression, fully-connected feedforward networks,
convolutional neural networks and the MEG kernel.
We show that end-to-end discriminative training of the GP adapter
outperforms a variety of baselines in terms of classification performance,
including models based only on GP mean interpolation,
or with GP regression trained separately using marginal likelihood.

%% file: irregular-time-series.tex
\section{Gaussian processes for sparse and irregularly-sampled time series}
\label{sec:gp}
Our focus in this paper is on time series classification in the
presence of sparse and irregular sampling.
In this problem, the data $\mathcal{D}$ contain
$N$ independent tuples consisting of a time series
$\mathcal{S}_i$ and a label $y_i$.
Thus, $\mathcal{D}=\{(\mathcal{S}_1,y_1),\dots,(\mathcal{S}_N,y_N)\}$.
Each time series $\mathcal{S}_i$ is represented as a list of
time points
$\tt_i=[t_{i1},\dots,t_{i|\mathcal{S}_i|}]^\top$,
and a list of corresponding values
$\vv_i=[v_{i1},\dots,v_{i|\mathcal{S}_i|}]^\top$.
We assume that each time series is observed over a common time interval
$[0,T]$.
However,
different time series are not necessarily observed at the same
time points (i.e.~$\tt_i\ne\tt_j$ in general).
This implies that the number of observations in different time series
is not necessary the same (i.e.~$|\mathcal{S}_i|\ne|\mathcal{S}_j|$ in general).
Furthermore, the time intervals between observation within a single time
series are not assumed to be uniform.

Learning in this setting is challenging because the data cases are not
naturally defined in a fixed-dimensional feature space due to
the irregular sampling.
This means that commonly used classifiers that take fixed-length feature
vectors as input are not applicable.
In addition, there can be substantial uncertainty
about the underlying temporal processes due to the sparsity
of observations.


To address these challenges, we build on ideas from the MEG kernel
\citep{li2015classification} by using GP regression
\citep{rasmussen2006gaussian} to provide an uncertainty-aware representation of
sparse and irregularly sampled time series.
We fix a set of reference time points $\xx=[x_1,\dots,x_d]^\top$
and represent a time series
$\mathcal{S}=(\tt,\vv)$ in terms of its posterior marginal distribution at
these time points.
We use GP regression with a zero-mean
GP prior and a covariance function $k(\cdot,\cdot)$ parameterized by
kernel hyperparameters $\eeta$.
Let $\sigma^2$ be the independent noise variance of the GP regression model.
The GP parameters are $\ttheta=(\eeta, \sigma^2)$.

Under this model, the marginal posterior GP at $\xx$ is Gaussian distributed
with the mean and covariance given by
\begin{align}
  \mmu &=
  \KK_{\xx,\tt}(\KK_{\tt,\tt} + \sigma^2 \I)^{-1}\vv,
  \label{eq:gp_mean} \\
  \SSigma &=
  \KK_{\xx,\xx} - \KK_{\xx,\tt}(\KK_{\tt,\tt} + \sigma^2 \I)^{-1}\KK_{\tt,\xx}
  \label{eq:gp_cov}
\end{align}
where $\KK_{\xx,\tt}$ denotes the
covariance matrix with $[\KK_{\xx,\tt}]_{ij} = k(x_i,t_j)$. We note that it
takes $\O(n^3 + nd)$ time to exactly compute the posterior mean $\mmu$, and
$\O(n^3 + n^2d + nd^2)$ time to exactly compute the full posterior covariance
matrix $\SSigma$, where $n=|\tt|$ and $d=|\xx|$.

%% file: gp-adapter.tex
\section{The GP adapter and uncertainty-aware time series classification}
\label{sec:gp-adapter}

In this section we describe our framework
for time series classification in the
presence of sparse and irregular sampling.
Our framework enables any black-box classifier learnable
by gradient-based methods
to be applied to the problem of classifying
sparse and irregularly sampled time series.
\subsection{Classification frameworks and the Gaussian process adapter}
\label{sec:frameworks}
In Section~\ref{sec:gp}
we described how we can represent a time series through the
marginal posterior it induces under a Gaussian process regression model at any
set of reference time points $\xx$. By fixing a common set of reference time
points $\xx$ for all time series in a data set, every time series
can be transformed into a common representation in the form of a multivariate
Gaussian $\mathcal{N}(\zz|\mmu,\SSigma;\ttheta)$
with $\zz$ being the random vector distributed according to
the posterior GP marginalized over the time points $\xx$.\footnote{
  The notation $\N(\mmu,\SSigma;\ttheta)$
  explicitly expresses that both $\mmu$ and $\SSigma$ are
  functions of the GP parameters $\ttheta$.
  Besides, they are also functions of $\mathcal{S}=(\tt,\vv)$
  as shown in \eqref{eq:gp_mean} and \eqref{eq:gp_cov}.
}
Here we assume that the GP parameters $\ttheta$ are shared across the
entire data set.

If the $\zz$ values were observed, we could simply apply a
black-box classifier.
A classifier can be generally defined by a mapping function $f(\zz;\ww)$
parameterized by $\ww$, associated with a loss function
$\ell(f(\zz;\ww),y)$ where $y$ is a label value
from the output space $\mathcal{Y}$.
However, in our case $\zz$ is a Gaussian random variable,
which means $\ell(f(\zz;\ww),y)$ is now itself a random variable
given a label $y$.
Therefore, we use the expectation
$\E_{\zz\sim\N(\mmu,\SSigma;\ttheta)}\big[\ell(f(\zz;\ww), y)\big]$
as the overall loss between the label $y$ and a time series $\mathcal{S}$
given its Gaussian representation $\N(\mmu,\SSigma;\ttheta)$.
The learning problem becomes minimizing the expected loss
over the entire data set:
\begin{equation}
  \ww^*,\ttheta^*
  =\argmin_{\ww,\ttheta}\sum_{i=1}^N
  \E_{\zz_i\sim\N(\mmu_i,\SSigma_i;\ttheta)}\big[\ell(f(\zz_i;\ww), y_i)\big].
  \label{eq:uac_train}
\end{equation}
Once we have the optimal parameters $\ww^*$ and $\ttheta^*$, we
can make predictions on unseen data.
In general, given an unseen time series $\mathcal{S}$ and its
Gaussian representation $\N(\mmu,\SSigma;\ttheta^*)$,
we can predict its label using \eqref{eq:uac_predict},
although in many cases this can be simplified into
a function of $f(\zz;\ww^*)$ with the expectation
taken on or inside of $f(\zz;\ww^*)$.
\begin{equation}
  y^* = \argmin_{y\in\mathcal{Y}}
  \E_{\zz\sim\N(\mmu,\SSigma;\ttheta^*)}\big[\ell(f(\zz;\ww^*), y)\big]
  \label{eq:uac_predict}
\end{equation}
We name the above approach the
\emph{Uncertainty-Aware Classification} (UAC) framework.
Importantly, this framework propagates the uncertainty in the GP posterior
induced by each time series all the way through to the loss function.
Besides, we call the transformation $\mathcal{S}\mapsto(\mmu,\SSigma)$
the \emph{Gaussian process adapter}, since
it provides a uniform representation to
connect the raw irregularly sampled time series data
to a black-box classifier.

Variations of the UAC framework can be derived by
taking the expectation at various position of $f(\zz;\ww)$
where $\zz\sim\N(\mmu,\SSigma;\ttheta)$.
Taking the expectation at an earlier stage simplifies the computation,
but the uncertainty information will be integrated out earlier
as well.\footnote{
For example, the loss of the expected output of the classifier
$\ell(\E_{\zz\sim\N(\mmu,\SSigma;\ttheta)}[f(\zz;\ww)], y)$.
}
In the extreme case, if the expectation is computed immediately followed by
the GP adapter transformation,
it is equivalent to using a plug-in estimate $\mmu$ for $\zz$
in the loss function,
$\ell(f(\E_{\zz\sim\N(\mmu,\SSigma;\ttheta)}[\zz];\ww), y)
=\ell(f(\mmu;\ww), y)$.
We refer to this as the IMPutation (IMP) framework.
The IMP framework discards the uncertainty information completely,
which further simplifies the computation.
This simplified variation may be useful
when the time series are more densely sampled,
where the uncertainty is less of a concern.


In practice, we can train the model using the
UAC objective \eqref{eq:uac_train} and predict instead by IMP.
In that case, the predictions would be deterministic and
can be computed efficiently without drawing samples from the posterior
GP as described later in Section~\ref{sec:gp_sampling}.

\subsection{Learning with the GP adapter}
In the previous section, we showed that the UAC framework
can be trained using \eqref{eq:uac_train}.
In this paper, we use stochastic gradient descent to scalably
optimize \eqref{eq:uac_train}
by updating the model using a single time series at a time,
although it can be easily modified for batch or mini-batch updates.
From now on, we will focus on the optimization problem
$\min_{\ww,\ttheta}
\E_{\zz\sim\N(\mmu,\SSigma;\ttheta)}\big[\ell(f(\zz;\ww), y)\big]$
where $\mmu,\SSigma$ are the output of the GP adapter given
a time series $\mathcal{S}=(\tt,\vv)$ and its label $y$.
For many classifiers, the expected loss
$\E_{\zz\sim\N(\mmu,\SSigma;\ttheta)}\big[\ell(f(\zz;\ww), y)\big]$
cannot be analytically computed.
In such cases, we use the Monte Carlo average to approximate the expected loss:
\begin{equation}
  \E_{\zz\sim\N(\mmu,\SSigma;\ttheta)}\big[\ell(f(\zz;\ww), y)\big]
  \approx\frac{1}{S}\sum_{s=1}^S \ell(f(\zz_s;\ww), y),
  \quad
  \text{where }\zz_s\sim\N(\mmu,\SSigma;\ttheta).
  \label{eq:exp_loss}
\end{equation}
To learn the parameters of both the classifier $\ww$ and
the Gaussian process regression model $\ttheta$ jointly
under the expected loss, we need to be able to compute
the gradient of the expectation given in \eqref{eq:exp_loss}.
To achieve this, we reparameterize the Gaussian random variable using the
identity $\zz=\mmu + \RR\xxi$ where $\xxi\sim\N(\0,\I)$ and
$\RR$ satisfies $\SSigma=\RR\RR^\top$ \citep{kingma2013auto}.
The gradients under this reparameterization are given below,
both of which can be
approximated using Monte Carlo sampling as in \eqref{eq:exp_loss}.
We will focus on efficiently computing the gradient shown in \eqref{eq:grad_gpa}
since we assume that the gradient of the base classifier $f(\zz;\ww)$ can
be computed efficiently.
\begin{align}
  \frac{\partial}{\partial\ww}
  \E_{\zz\sim\N(\mmu,\SSigma;\ttheta)}
  \big[\ell(f(\zz;\ww), y)\big]
  &=
    \E_{\xxi\sim\N(\0,\I)}
    \left[\frac{\partial}{\partial\ww}
    \ell(f(\zz;\ww), y)\right]
    \label{eq:grad_ietwork}
  \\
  \frac{\partial}{\partial\ttheta}
  \E_{\zz\sim\N(\mmu,\SSigma;\ttheta)}
  \big[\ell(f(\zz;\ww), y)\big]
  &=
    \E_{\xxi\sim\N(\0,\I)}
    \left[
      \sum_i
      \frac{\partial\ell(f(\zz;\ww), y)}{\partial z_i}
      \frac{\partial z_i}{\partial\ttheta}
    \right]
    \label{eq:grad_gpa}
\end{align}
There are several choices for $\RR$ that satisfy $\SSigma=\RR\RR^\top$.
One common choice of $\RR$ is the Cholesky factor,
a lower triangular matrix, which can be computed using
Cholesky decomposition  in $\O(d^3)$ for
a $d\times d$ covariance matrix $\SSigma$ \citep{golub2012matrix}.
We instead use the symmetric matrix square root $\RR=\SSigma^{\half}$.
We will show that this particular choice of $\RR$
leads to an efficient and scalable
approximation algorithm in Section~\ref{sec:lanczos}.

%% file: fast_sampling.tex
\section{Fast sampling from posterior Gaussian processes}
\label{sec:gp_sampling}
The computation required by the GP adapter is dominated by the time needed to
draw samples from the marginal GP posterior using
$\zz = \mmu + \SSigma^{\half}\xxi$.
In Section~\ref{sec:gp}
we noted that the time complexity of exactly computing the
posterior mean $\mmu$ and covariance $\SSigma$ is $\O(n^3 + nd)$ and
$\O(n^3 + n^2d + nd^2)$, respectively.
Once we have both $\mmu$ and $\SSigma$ we still
need to compute the square root of $\SSigma$, which requires an additional
$\O(d^3)$ time to compute exactly.
In this section, we show how to efficiently generate samples of $\zz$.

%
\subsection{Structured kernel interpolation for approximating GP posterior means}
The main idea of the structured kernel interpolation (SKI) framework
recently proposed by \citet{wilson2015kernel} is to approximate
a stationary kernel matrix $\KK_{\aa,\bb}$
by the approximate kernel $\widetilde\KK_{\aa,\bb}$
defined below where $\uu=[u_1,\dots,u_m]^\top$ is a collection of
evenly-spaced inducing points.
\begin{equation}
  \KK_{\aa,\bb} \approx
  \widetilde\KK_{\aa,\bb} = \WW_{\aa}\KK_{\uu,\uu}\WW_{\bb}^\top.
  \label{eq:ski}
\end{equation}
Letting $p=|\aa|$ and $q=|\bb|$,
$\WW_{\aa}\in\R^{p\times m}$
is a sparse interpolation matrix where each row contains
only a small number of non-zero entries.
%
We use local cubic convolution interpolation (cubic interpolation for short)
\citep{keys1981cubic} as suggested in \citet{wilson2015kernel}.
Each row of the interpolation matrices $\WW_{\aa},\WW_{\bb}$
has at most four non-zero entries.
\citet{wilson2015kernel} showed that when the kernel is locally smooth (under
the resolution of $\uu$), cubic interpolation results in accurate
approximation.
This can be justified as follows:
with cubic interpolation, the SKI kernel is essentially
the two-dimensional cubic interpolation of $\KK_{\aa,\bb}$
using the exact regularly spaced samples stored in $\KK_{\uu,\uu}$,
which corresponds to classical bicubic convolution.
In fact,
we can show that $\widetilde\KK_{\aa,\bb}$ asymptotically converges to
$\KK_{\aa,\bb}$ as $m$ increases
by following the derivation in \citet{keys1981cubic}.

Plugging the SKI kernel into \eqref{eq:gp_mean},
the posterior GP mean evaluated at $\xx$ can be approximated by
\begin{equation}
  \mmu
  =
  \KK_{\xx,\tt}\left(\KK_{\tt,\tt} + \sigma^2 \I\right)^{-1} \vv
  \approx
  \WW_{\xx}\KK_{\uu,\uu}\WW_{\tt}^\top\left(
  \WW_{\tt} \KK_{\uu,\uu}^{-1} \WW_{\tt}^\top + \sigma^2 \I\right)^{-1} \vv.
  \label{eq:ski_mean}
\end{equation}
The inducing points $\uu$ are chosen to be evenly-spaced because
$\KK_{\uu,\uu}$ forms a symmetric Toeplitz matrix
under a stationary covariance function.
A symmetric Toeplitz matrix can be embedded into a circulant matrix
to perform matrix vector multiplication using fast Fourier transforms
\citep{golub2012matrix}.

Further, one can use the conjugate gradient method to solve for
$(\WW_{\tt} \KK_{\uu,\uu}^{-1} \WW_{\tt}^\top + \sigma^2 \I)^{-1}\vv$
which only involves computing the matrix-vector product
$(\WW_{\tt} \KK_{\uu,\uu}^{-1} \WW_{\tt}^\top + \sigma^2 \I)\vv$.
In practice, the conjugate gradient method converges within only a few
iterations.
Therefore, approximating the posterior mean $\mmu$ using SKI takes only
$\O(n + d + m\log m)$ time to compute.
In addition, since a symmetric Toeplitz matrix $\KK_{\uu,\uu}$
can be uniquely characterized by its first column,
and $\WW_{\tt}$ can be stored as a sparse matrix,
approximating $\mmu$ requires only $\O(n + d + m)$ space.

%
\subsection{The Lanczos method for covariance square root-vector products}
\label{sec:lanczos}
With the SKI techniques,
although we can efficiently approximate the posterior mean $\mmu$,
computing $\SSigma^{\half}\xxi$ is still challenging.
If computed exactly, it takes $\O(n^3 + n^2d + nd^2)$ time to compute
$\SSigma$ and $\O(d^3)$ time to take the square root.
To overcome the bottleneck,
we apply the SKI kernel to the Lanczos method,
one of the Krylov subspace approximation methods,
to speed up the computation of $\SSigma^{\half}\xxi$ as shown in
Algorithm~\ref{algo:lanczos}.
The advantage of the Lanczos method is that neither $\SSigma$
nor $\SSigma^{\half}$ needs to be computed explicitly.
Like the conjugate gradient method, another example of
the Krylov subspace method, it only requires the computation of
 matrix-vector products with $\SSigma$ as the matrix.

The idea of the Lanczos method is to approximate
$\SSigma^{\half}\xxi$ in the Krylov subspace
  $K_k(\SSigma,\xxi)=\operatorname{span}
  \{\xxi,\SSigma\xxi,\dots,\SSigma^{k-1}\xxi\}$.
The iteration in Algorithm~\ref{algo:lanczos},
usually referred to the Lanczos process,
essentially performs the Gram-Schmidt process to transform the basis $\{\xxi,\SSigma\xxi,\dots,\SSigma^{k-1}\xxi\}$
into an orthonormal basis $\{\dd_1,\dots,\dd_k\}$ for the subspace
$K_k(\SSigma,\xxi)$.

The optimal approximation of $\SSigma^{\half}\xxi$ in
the Krylov subspace $K_k(\SSigma,\xxi)$
that minimizes the $\ell^2$-norm of the error is
the orthogonal projection of $\SSigma^{\half}\xxi$ onto
$K_k(\SSigma,\xxi)$
as $\yy^* = \DD\DD^\top\SSigma^{\half}\xxi$.
Since we choose $\dd_1=\xxi/\|\xxi\|$,
the optimal projection can be written as
$\yy^* = \|\xxi\|\DD\DD^\top\SSigma^{\half}\DD\ee_1$
where $\ee_1 = [1,0,\dots,0]^\top$
is the first column of the identify matrix.

One can show that the tridiagonal matrix $\HH$ defined in
Algorithm~\ref{algo:lanczos} satisfies $\DD^\top\SSigma\DD=\HH$
\citep{saad2003iterative}.
Also, we have $\DD^\top\SSigma^{\half}\DD \approx (\DD^\top\SSigma\DD)^{\half}$
since the eigenvalues of $\HH$ approximate the extremal eigenvalues
of $\SSigma$ \citep{saad1980rates}.
Therefore we have
  $\yy^* = \|\xxi\|\DD\DD^\top\SSigma^{\half}\DD\ee_1
  \approx \|\xxi\|\DD\HH^{\half}\ee_1$.

The error bound of the Lanczos method is analyzed in
\citet{ilic2009restarted}.
Alternatively one can show that
the Lanczos approximation converges superlinearly
\citep{parlett1980symmetric}.
In practice, for a $d\times d$ covariance matrix $\SSigma$, the approximation
is sufficient for our sampling purpose with $k\ll d$.
As $\HH$ is now a $k\times k$ matrix, we can use any standard method
to compute its square root in $\O(k^3)$ time \citep{bjorck1983schur},
which is considered $\O(1)$ when $k$ is chosen to be a small constant.
Now the computation of the Lanczos method for approximating
$\SSigma^{\half}\xxi$ is dominated by the
matrix-vector product $\SSigma\dd$ during the Lanczos process.


\begin{figure}[t]
  \begin{tikzpicture}
    \node[inner sep=0, outer sep=0] {
      \begin{varwidth}{\linewidth}
      \begin{algorithm}[H]
        \DontPrintSemicolon
        \KwIn{covariance matrix $\SSigma$,
          dimension of the Krylov subspace $k$, random vector $\xxi$}
        $\beta_1 = 0$ and $\dd_0 = \0$\;
        $\dd_1 = \xxi/\|\xxi\|$\;
        \For{$j=1$ \KwTo $k$}{
          $\dd=\SSigma\dd_j - \beta_j\dd_{j-1}$\;
          $\alpha_j = \dd_j^\top\dd$\;
          $\dd = \dd - \alpha_j\dd_j$\;
          $\beta_{j+1}=\|\dd\|$\;
          $\dd_{j+1}=\dd/\beta_{j+1}$\;
        }
        $\DD=[\dd_1,\dots,\dd_k]$\;
        $\HH=\operatorname{tridiagonal}(\bbeta, \aalpha, \bbeta)$\;
        \Return{$\|\xxi\|\DD\HH^{\half}\ee_1$}
        \tcp*[r]{$\ee_1=[1,0,\dots,0]^\top$}
        \caption{Lanczos method for approximating $\SSigma^{\half}\xxi$}
        \label{algo:lanczos}
      \end{algorithm}
    \end{varwidth}
    };
    \node[inner sep=0, outer sep=0, anchor=west] at (-1.6,-.8) {
      \scalebox{1}{
        \begin{varwidth}{\linewidth}
          \begin{empheq}[box=\fbox]{align*}
            \HH=\operatorname{tridiagonal}(\bbeta, \aalpha, \bbeta)=
            \begin{bmatrix}
              \alpha_1 & \beta_2 & & & \\[.1em]
              \beta_2 & \alpha_2 & \beta_3 & & \\[-.5em]
              & \beta_3 & \alpha_3 & \ddots & \\[-.3em]
              & & \ddots & \ddots & \beta_k \\
              & & & \beta_k & \alpha_k \\
            \end{bmatrix}
          \end{empheq}
        \end{varwidth}
      }
    };
  \end{tikzpicture}
\end{figure}

Here we apply the SKI kernel trick again to efficiently
approximate $\SSigma\dd$ by
\begin{equation}
  \SSigma\dd
  \approx
  \WW_{\xx}\KK_{\uu,\uu}\WW_{\xx}^\top\dd
  -\WW_{\xx}\KK_{\uu,\uu}\WW_{\tt}^\top
  \left(\WW_{\tt}\KK_{\uu,\uu}\WW_{\tt}^\top + \sigma^2\I\right)^{-1}
  \WW_{\tt}\KK_{\uu,\uu}\WW_{\xx}^\top\dd.
  \label{eq:ski_cov}
\end{equation}
Similar to the posterior mean, $\SSigma\dd$ can be
approximated in $\O(n + d + m\log m)$ time and linear space.
Therefore, for $k=\O(1)$ basis vectors, the entire Algorithm~\ref{algo:lanczos}
takes $\O(n + d + m\log m)$ time and $\O(n + d + m)$ space,
which is also the complexity to draw a sample from the posterior GP.


To reduce the variance when estimating the expected loss \eqref{eq:exp_loss},
we can draw multiple samples from the posterior GP:
$\{\SSigma^{\half}\xxi_s\}_{s=1,\dots,S}$ where $\xxi_s\sim\N(\0,\I)$.
Since all of the samples are associated with the same covariance matrix $\SSigma$,
we can use the block Lanczos process \citep{golub1977block},
an extension to the single-vector Lanczos method presented in
Algorithm~\ref{algo:lanczos},
to simultaneously approximate $\SSigma^{\half}\XXi$
for all $S$ random vectors $\XXi=[\xxi_1,\dots,\xxi_S]$.
Similarly, during the block Lanczos process,
we use the block conjugate gradient method
\citep{feng1995block,dubrulle2001retooling} to
simultaneously solve the linear equation
$(\WW_{\tt}\KK_{\uu,\uu}\WW_{\tt}^\top + \sigma^2\I)^{-1}\aalpha$
for multiple $\aalpha$.


%% file: end_to_end.tex
\section{End-to-end learning with the GP adapter}
\label{sec:backprop}

The most common way to train GP parameters is through maximizing
the marginal likelihood \citep{rasmussen2006gaussian}
\begin{equation}
  \log p(\vv|\tt,\ttheta) =
  -\frac{1}{2}\vv^\top\left(\KK_{\tt,\tt} + \sigma^2\I\right)^{-1}\vv
  -\frac{1}{2}\log\left|\KK_{\tt,\tt} + \sigma^2\I\right|
  -\frac{n}{2}\log 2\pi.
\end{equation}
If we follow this criterion, training the UAC framework becomes
a two-stage procedure: first we learn GP parameters by maximizing
the marginal likelihood. We then compute $\mmu$ and $\SSigma$
given each time series $\mathcal{S}$ and the learned GP parameters $\ttheta^*$.
Both $\mmu$ and $\SSigma$
are then fixed and used to train the classifier using \eqref{eq:grad_ietwork}.

In this section, we describe how to instead train the GP parameters
discriminatively end-to-end using backpropagation.
As mentioned in Section~\ref{sec:gp-adapter},
we train the UAC framework by jointly optimizing
the GP parameters $\ttheta$ and the parameters of the classifier $\ww$
according to \eqref{eq:grad_ietwork} and \eqref{eq:grad_gpa}.

The most challenging part in \eqref{eq:grad_gpa}
is to compute
$\partial\zz=\partial\mmu+\partial(\SSigma^{\half}\xxi)$.\footnote{
For brevity, we drop $1/\partial\ttheta$ from the gradient notation
in this section.}
For $\partial\mmu$,
we can derive the gradient of the approximating
posterior mean \eqref{eq:ski_mean} as given in Appendix~\ref{sec:grad}.
Note that the gradient $\partial\mmu$ can be approximated efficiently
by repeatedly applying fast Fourier transforms
and the conjugate gradient method in the same time and space
complexity as computing \eqref{eq:ski_mean}.

On the other hand,
$\partial(\SSigma^{\half}\xxi)$ can be approximated
by backpropagating through the Lanczos method described in
Algorithm~\ref{algo:lanczos}.
To carry out backpropagation, all operations in the Lanczos method
must be differentiable.
For the approximation of $\SSigma\dd$ during the Lanczos process,
we can similarly compute the gradient of \eqref{eq:ski_cov}
efficiently using the SKI techniques as in computing $\partial\mmu$
(see Appendix~\ref{sec:grad}).

The gradient $\partial\HH^{\half}$
for the last step of Algorithm~\ref{algo:lanczos}
can be derived as follows.
From $\HH = \HH^{\half}\HH^{\half}$,
we have $\partial\HH =
(\partial\HH^{\half})\HH^{\half} + \HH^{\half}(\partial\HH^{\half})$.
This is known as the Sylvester equation, which has the form of
$\bm{A}\bm{X} + \bm{X}\bm{B} = \bm{C}$
where $\bm{A}, \bm{B}, \bm{C}$ are matrices and $\bm{X}$ is
the unknown matrix to solve for.
We can compute the gradient $\partial\HH^{\half}$ by
solving the Sylvester equation using
the Bartels-Stewart algorithm \citep{bartels1972solution}
in $\O(k^3)$ time for a $k\times k$ matrix $\HH$,
which is considered $\O(1)$ for a small constant $k$.

Overall, training the GP adapter using stochastic optimization
with the aforementioned approach
takes $\O(n + d + m\log m)$ time and $\O(n + d + m)$ space
for $m$ inducing points, $n$ observations in the time series,
and $d$ features generated by the GP adapter.


%% file: related.tex
\section{Related work}
\label{sec:related}

The recently proposed
mixtures of expected Gaussian kernels (MEG) \citep{li2015classification}
for classification of irregular time series is probably the closest work
to ours.
The random feature representation of the MEG kernel is in the form of
$\sqrt{2/m}\;\E_{\zz\sim\N(\mmu,\SSigma)}\left[\cos(\ww_i^\top\zz+b_i)\right]$,
which the algorithm described in Section~\ref{sec:gp_sampling} can
be applied to directly.
However, by exploiting the spectral property of Gaussian kernels,
the expected random feature of the MEG kernel is shown to
be analytically computable by
$\sqrt{2/m}\,\exp(-\ww_i^\top\SSigma\ww_i/2)\cos(\ww_i^\top\mmu+b_i)$.
With the SKI techniques, we can efficiently approximate
both $\ww_i^\top\SSigma\ww_i$ and $\ww_i^\top\mmu$
in the same time and space complexity as the GP adapter.
Moreover, the random features of the MEG kernel can be viewed
as a stochastic layer in the classification network,
with no trainable parameters.
All $\{\ww_i,b_i\}_{i=1,\dots,m}$ are randomly initialized
once in the beginning and associated with the output of the
GP adapter in a nonlinear way described above.

Moreover, the MEG kernel classification is originally a two-stage method:
one first estimates the GP parameters by maximizing the marginal
likelihood and then uses the optimized GP parameters to compute
the MEG kernel for classification.
Since the random feature is differentiable, with the approximation
of $\partial\mmu$ and $\partial(\SSigma\dd)$ described in
Section~\ref{sec:backprop},
we can form a similar classification network that can be efficiently trained
end-to-end using the GP adapter.
In Section~\ref{sec:data_exp},
we will show that training the MEG kernel end-to-end
leads to better classification performance.


%% file: experiments.tex
\section{Experiments}
\label{sec:experiments}

In this section, we present experiments and results exploring several facets of the GP adapter framework including the quality of the approximations and the classification performance of the framework when combined with different base classifiers.

\subsection{Quality of GP sampling approximations}

\begin{figure}[tb]
  \centering
  \hspace{-.2in}
  \includegraphics[height=1.35in]{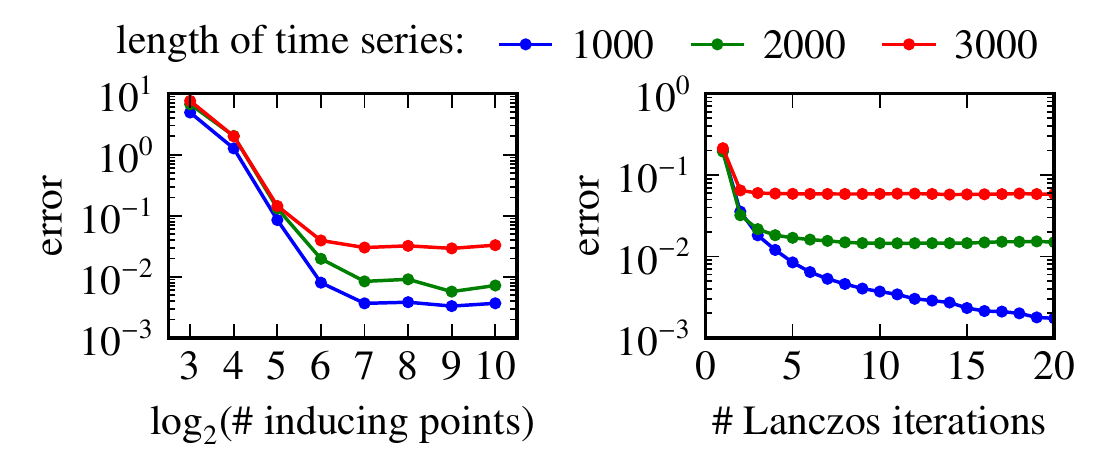}
  \includegraphics[height=1.46in]{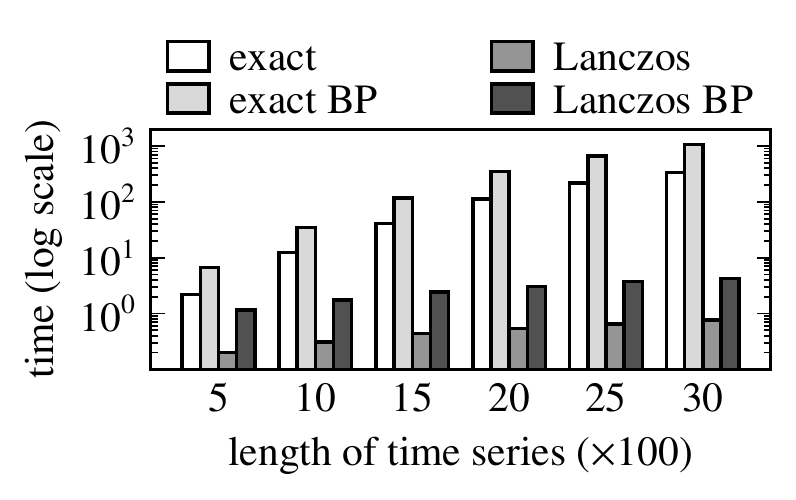}
  \caption{
      Left: Sample approximation error versus the number of inducing points.
      Middle: Sample approximation error versus
      the number of Lanczos iterations.
      Right: Running time comparisons (in seconds).
      BP denotes computing the gradient of the sample using backpropagation.
  }
  \label{fig:cmp}
\end{figure}

The key to scalable learning with the GP adapter relies on both fast
and accurate approximation for drawing samples from the posterior GP.
To assess the approximation quality,
we first generate a synthetic sparse and irregularly-sampled time series
$\mathcal{S}$
by sampling from a zero-mean Gaussian process at random time points.
We use the squared exponential kernel
$k(t_i,t_j)=a\exp(-b(t_i-t_j)^2)$ with randomly chosen hyperparameters.
We then infer $\mmu$ and $\SSigma$ at some
reference $\xx$ given $\mathcal{S}$.
Let $\widetilde{\zz}$ denote our approximation of
$\zz=\mmu + \SSigma^{\half}\xxi$.
In this experiment, we set the output size $\zz$ to be $|\mathcal{S}|$,
that is, $d=n$.
We evaluate the approximation quality by assessing
the error $\|\widetilde\zz - \zz\|$ computed with a fixed
random vector $\xxi$.

The leftmost plot in Figure~\ref{fig:cmp}
shows the approximation error under different numbers of inducing points $m$
with $k=10$ Lanczos iterations.
The middle plot compares the approximation error as the number
of Lanczos iterations $k$ varies, with $m=256$ inducing points.
These two plots show that the approximation error drops as more inducing points
and Lanczos iterations are used.
In both plots, the three lines correspond to
different sizes for $\zz$:
1000 (bottom line), 2000 (middle line), 3000 (top line).
The separation between the curves is due to the fact that
the errors are compared under the same number of inducing points.
Longer time series leads to lower resolution of the inducing points
and hence the higher approximation error.

Note that the approximation error comes from both
the cubic interpolation and the Lanczos method.
Therefore, to achieve a certain normalized approximation error across different
data sizes, we should simultaneously use more inducing points and
Lanczos iterations as the data grows.
In practice, we find that $k \ge 3$ is sufficient for
estimating the expected loss for classification.

The rightmost plot in Figure~\ref{fig:cmp} compares the time to draw
a sample using exact computation versus the approximation method
described in Section~\ref{sec:gp_sampling} (exact and Lanczos
in the figure).
We also compare the time to compute the gradient with respect to
the GP parameters by both the exact method and the proposed approximation
(exact BP and Lanczos BP in the figure)
because this is the actual computation carried out during training.
In this part of the experiment, we use $k=10$ and $m=256$.
The plot shows that Lanczos approximation with the SKI kernel
yields speed-ups of between 1 and 3 orders of magnitude.
Interestingly, for the exact approach,
the time for computing the gradient roughly doubles the time of drawing samples.
(Note that time is plotted in log scale.)
This is because
computing gradients requires both forward and backward propagation,
whereas drawing samples corresponds to only the forward pass.
Both the forward and backward passes take roughly the same computation
in the exact case.
However, the gap is relatively larger for the approximation approach
due to the recursive relationship of the variables in the Lanczos process.
In particular, $\dd_j$ is defined recursively
in terms of all of $\dd_1,\dots,\dd_{j-1}$, which
makes the backpropagation computation more complicated
than the forward pass.


\subsection{Classification with GP adapter}
\label{sec:data_exp}

\begin{table}[t]
  \caption{Comparison of classification accuracy (in percent).
    IMP and UAC refer to the loss functions for training described in
    Section~\ref{sec:frameworks}, and
    we use IMP predictions throughout.
    Although not belonging to the UAC framework,
    we put the MEG kernel in UAC
    since it is also uncertainty-aware.
    }
  \label{tab:accuracy}
  \centering
  \newcolumntype{C}{>{\centering\arraybackslash}m{.68in}}
  \begin{tabular}{lcCCCC}
    \toprule
    & & LogReg & MLP & ConvNet & MEG kernel \\
    \midrule
    \multirow{2}{*}{Marginal likelihood}
    & IMP
    & 77.90
    & 85.49
    & 87.61
    & --
    \\
    & UAC
    & 78.23
    & 87.05
    & 88.17
    & 84.82
    \\
    \cmidrule{1-6}
    \multirow{2}{*}{End-to-end}
    & IMP
    & 79.12
    & 86.49
    & 89.84
    & --
    \\
    & UAC
    & \textbf{79.24}
    & \textbf{87.95}
    & \textbf{91.41}
    & \textbf{86.61}
    \\
    \bottomrule
  \end{tabular}
\end{table}

In this section, we evaluate the performance of classifying
sparse and irregularly-sampled time series using the UAC framework.
We test the framework on the uWave data set,\footnote{
  The data set \texttt{UWaveGestureLibraryAll}
  is available at \url{http://timeseriesclassification.com}.}
a collection of gesture samples
categorized into eight gesture patterns \citep{liu2009uwave}.
The data set has been split into 3582 training instances and 896 test instances.
Each time series contains 945 fully observed samples.
Following the data preparation procedure in the MEG kernel work
\citep{li2015classification},
we randomly sample 10\% of the observations from each time series to simulate
the sparse and irregular sampling scenario.
In this experiment, we use the squared exponential covariance function
$k(t_i,t_j) = a\exp(-b(t_i-t_j)^2)$ for $a,b>0$.
Together with the independent noise parameter $\sigma^2 > 0$,
the GP parameters are $\{a, b, \sigma^2\}$.
To bypass the positive constraints on the GP parameters,
we reparameterize them by $\{\alpha, \beta, \gamma\}$
such that $a = e^\alpha$, $b=e^\beta$, and $\sigma^2=e^\gamma$.

To demonstrate that the GP adapter is capable of
working with various classifiers,
we use the UAC framework to train three different classifiers:
a multi-class logistic regression
(LogReg), a fully-connected feedforward network (MLP),
and a convolutional neural network (ConvNet).
The detailed architecture of each model is described in Appendix~\ref{sec:arch}.


We use $m=256$ inducing points,
$d=254$ features output by the GP adapter,
$k=5$ Lanczos iterations, and $S=10$ samples.
We split the training set into two partitions:
$70\%$ for training and $30\%$ for validation.
We jointly train the classifier with the GP adapter using
stochastic gradient descent with Nesterov momentum.
We apply early stopping based on the validation set.
We also compare to classification with the MEG kernel
implemented using our GP adapter as described in Section~\ref{sec:related}.
We use $1000$ random features trained with multi-class logistic regression.

Table~\ref{tab:accuracy} shows that
among all three classifiers, training GP parameters discriminatively
always leads to better accuracy than maximizing the marginal likelihood.
This claim also holds for the results using the MEG kernel.
Further, taking the uncertainty into account by
sampling from the posterior GP always outperforms
training using only the posterior means.
Finally, we can also see that the classification accuracy improves as
the model gets deeper.

%% file: conclusion.tex
\section{Conclusions and future work}

We have presented a general framework for classifying
sparse and irregularly-sampled time series and have shown how
to scale up the required computations using a new approach to
generating approximate samples. We have validated the
approximation quality, the computational
speed-ups, and the benefit of the proposed approach
relative to existing baselines.

There are many promising directions for future work including
investigating more complicated covariance functions like
the spectral mixture kernel \citep{wilson2013gaussian},
different classifiers including the encoder
LSTM \citep{sutskever2014sequence}, and
extending the framework to multi-dimensional time series and
GPs with multi-dimensional index sets (e.g., for spatial data).
Lastly, the GP adapter can also be applied to other problems
such as dimensionality reduction
by combining it with an autoencoder.

%% file: acknowledgements.tex
\subsection*{Acknowledgements}

This work was supported by the National Science Foundation under
Grant No.\ 1350522.

%% file: gradients.tex
\section{Gradients for GP approximation}
\label{sec:grad}
\subsection{Gradients of the approximate posterior GP covariance-vector
product}
Throughout we denote the independent noise
variance $\sigma^2$ as $\rho$ for clarity.
Let $\widetilde\SSigma$ be the approximate posterior covariance
derived by the SKI kernel,
and $\theta$ be one of the GP hyperparameters.
For any vector $\dd$, the gradient
$\partial\widetilde\SSigma\dd/\partial\theta$ is given below.
Note that during the Lanczos process, $\dd$ is a function of $\theta$,
which should be properly handled in backpropagation.
\begin{align*}
  &
  \frac{\partial}{\partial\theta}
  \left\{
  \WW_{\xx}\KK_{\uu,\uu}\WW_{\xx}^\top\dd -
  \WW_{\xx}\KK_{\uu,\uu}\WW_{\tt}^\top
  \left(\WW_{\tt}\KK_{\uu,\uu}\WW_{\tt}^\top + \rho\I\right)^{-1}
  \WW_{\tt}\KK_{\uu,\uu}\WW_{\xx}^\top\dd
  \right\}
  \\
  &=
  \begin{aligned}[t]
  &
  \WW_{\xx}
  \frac{\partial\KK_{\uu,\uu}}{\partial\theta}
  \WW_{\xx}^\top\dd \\
  -\; &
  \WW_{\xx}
  \frac{\partial\KK_{\uu,\uu}}{\partial\theta}
  \WW_{\tt}^\top
  \left(\WW_{\tt}\KK_{\uu,\uu}\WW_{\tt}^\top + \rho\I\right)^{-1}
  \WW_{\tt}\KK_{\uu,\uu}\WW_{\xx}^\top\dd \\
  -\; &
  \WW_{\xx}\KK_{\uu,\uu}\WW_{\tt}^\top
  \left(\WW_{\tt}\KK_{\uu,\uu}\WW_{\tt}^\top + \rho\I\right)^{-1}
  \WW_{\tt}
  \frac{\partial\KK_{\uu,\uu}}{\partial\theta}
  \WW_{\xx}^\top\dd \\
  +\; &
  \WW_{\xx}\KK_{\uu,\uu}\WW_{\tt}^\top
  \left(\WW_{\tt}\KK_{\uu,\uu}\WW_{\tt}^\top + \rho\I\right)^{-1}
  \WW_{\tt}
  \frac{\partial\KK_{\uu,\uu}}{\partial\theta}
  \WW_{\tt}^\top \\
  &
  \left(\WW_{\tt}\KK_{\uu,\uu}\WW_{\tt}^\top + \rho\I\right)^{-1}
  \WW_{\tt}\KK_{\uu,\uu}\WW_{\xx}^\top\dd.
  \end{aligned}
\end{align*}

To reduce redundant computations, we introduce the following variables:
\begin{align*}
\aalpha &=
  \WW_{\xx}^\top\dd, \\
\bbeta &=
  \frac{\partial\KK_{\uu,\uu}}{\partial\theta}\aalpha, \\
\ggamma &=
  \KK_{\uu,\uu}\aalpha, \\
\ddelta &=
  \left(\WW_{\tt}\KK_{\uu,\uu}\WW_{\tt}^\top + \rho\I\right)^{-1}
  \WW_{\tt}\ggamma, \\
\zzeta &=
  \frac{\partial\KK_{\uu,\uu}}{\partial\theta}
  \WW_{\tt}^\top\ddelta, \\
\eeta &=
  \KK_{\uu,\uu}\WW_{\tt}^\top
  \left(\WW_{\tt}\KK_{\uu,\uu}\WW_{\tt}^\top + \rho\I\right)^{-1}
  \WW_{\tt}\left(\zzeta - \bbeta\right).
\end{align*}

The gradient with respect to $\theta$ is therefore
$\partial\widetilde\SSigma\dd/\partial\theta=
\WW_{\xx}\left(\bbeta - \zzeta + \eeta\right)$.

The gradient
$\partial\widetilde\SSigma\dd/\partial\rho$
with respect to the noise variance $\rho$ is given by
\begin{align*}
  &
  \frac{\partial}{\partial\rho}
  \left\{
  \WW_{\xx}\KK_{\uu,\uu}\WW_{\xx}^\top -
  \WW_{\xx}\KK_{\uu,\uu}\WW_{\tt}^\top
  \left(\WW_{\tt}\KK_{\uu,\uu}\WW_{\tt}^\top + \rho\I\right)^{-1}
  \WW_{\tt}\KK_{\uu,\uu}\WW_{\xx}^\top\dd
  \right\}
  \\
  &=
  \WW_{\xx}\KK_{\uu,\uu}\WW_{\tt}^\top
  \left(\WW_{\tt}\KK_{\uu,\uu}\WW_{\tt}^\top + \rho\I\right)^{-1}
  \left(\WW_{\tt}\KK_{\uu,\uu}\WW_{\tt}^\top + \rho\I\right)^{-1}
  \WW_{\tt}\KK_{\uu,\uu}\WW_{\xx}^\top\dd \\
  &=
  \WW_{\xx}\KK_{\uu,\uu}\WW_{\tt}^\top
  \left(\WW_{\tt}\KK_{\uu,\uu}\WW_{\tt}^\top + \rho\I\right)^{-1}
  \ddelta.
\end{align*}

\subsection{Gradients of the approximate posterior GP mean}
Let $\widetilde\mmu$ denote the approximate posterior mean derived by
the SKI kernel.
The gradient
$\partial\widetilde\mmu/\partial\theta$
with respect to the GP hyperparameter $\theta$ is given by
\begin{align*}
  &
  \frac{\partial}{\partial\theta}
  \WW_{\xx}\KK_{\uu,\uu}\WW_{\tt}^\top
  \left(\WW_{\tt}\KK_{\uu,\uu}\WW_{\tt}^\top + \rho\I\right)^{-1}
  \vv \\
  &=
  \begin{aligned}[t]
  &
  \WW_{\xx}
  \frac{\partial\KK_{\uu,\uu}}{\partial\theta}
  \WW_{\tt}^\top
  \left(\WW_{\tt}\KK_{\uu,\uu}\WW_{\tt}^\top + \rho\I\right)^{-1}
  \vv \\
  -\; &
  \WW_{\xx}\KK_{\uu,\uu}\WW_{\tt}^\top
  \left(\WW_{\tt}\KK_{\uu,\uu}\WW_{\tt}^\top + \rho\I\right)^{-1}
  \WW_{\tt}
  \frac{\partial\KK_{\uu,\uu}}{\partial\theta}
  \WW_{\tt}^\top
  \left(\WW_{\tt}\KK_{\uu,\uu}\WW_{\tt}^\top + \rho\I\right)^{-1}
  \vv.
  \end{aligned}
\end{align*}

To reduce redundant computations, we introduce the following variables:
\begin{align*}
\aalpha &=
  \left(\WW_{\tt}\KK_{\uu,\uu}\WW_{\tt}^\top + \rho\I\right)^{-1}
  \vv, \\
\bbeta &=
  \frac{\partial\KK_{\uu,\uu}}{\partial\theta}
  \WW_{\tt}^\top\aalpha, \\
\ggamma &=
  \KK_{\uu,\uu}\WW_{\tt}^\top
  \left(\WW_{\tt}\KK_{\uu,\uu}\WW_{\tt}^\top + \rho\I\right)^{-1}
  \WW_{\tt}\bbeta.
\end{align*}

The gradient with respect to $\theta$ is therefore
$\partial\widetilde\mmu/\partial\theta=
\WW_{\xx}\left(\bbeta - \ggamma\right)$.

The gradient
$\partial\widetilde\mmu/\partial\rho$
with respect to the noise variance $\rho$ is given by
\begin{align*}
  &
  \frac{\partial}{\partial\rho}
  \WW_{\xx}\KK_{\uu,\uu}\WW_{\tt}^\top
  \left(\WW_{\tt}\KK_{\uu,\uu}\WW_{\tt}^\top + \rho\I\right)^{-1}
  \vv \\
  &=
  -\WW_{\xx}\KK_{\uu,\uu}\WW_{\tt}^\top
  \left(\WW_{\tt}\KK_{\uu,\uu}\WW_{\tt}^\top + \rho\I\right)^{-1}
  \left(\WW_{\tt}\KK_{\uu,\uu}\WW_{\tt}^\top + \rho\I\right)^{-1}
  \vv \\
  &=
  -\WW_{\xx}\KK_{\uu,\uu}\WW_{\tt}^\top
  \left(\WW_{\tt}\KK_{\uu,\uu}\WW_{\tt}^\top + \rho\I\right)^{-1}
  \aalpha.
\end{align*}

%% file: cubic-interp.tex
\section{Cubic interpolation in backpropagation}
\label{sec:interp}
The choice of cubic convolution interpolation proposed by \citet{keys1981cubic}
is preferable over other interpolation methods
such as spline interpolation when training the GP parameters.
If spline interpolation is used to construct the SKI kernel
$\widetilde\KK_{\aa,\bb}$, 
the interpolation matrix $\WW_{\aa}$ depends not only on $\aa$ and $\uu$
but also on the kernel $\KK_{\uu,\uu}$, which depends on
the GP parameters $\ttheta$.
As a result, the gradient $\partial\WW_{\aa}/\partial\ttheta$
needs to be computed and thus introduces a huge overhead in backpropagation.
On the other hand, the interpolation matrix based on
the cubic convolution interpolation depends only on $\aa$ and $\uu$, which
are fixed once the data are given.
Therefore, with cubic convolution interpolation,
both $\WW_{\aa}$ and $\WW_{\bb}$ are constant matrices
throughout the entire training process.

%% file: arch.tex
\section{Architectures used in the experiment}
\label{sec:arch}
The architecture of each classifier compared in Section~\ref{sec:data_exp} are
described as follows.
The fully-connected network consists of two fully-connected layers,
each of which contains $256$ units.
The convolutional network contains a total of five layers:
the first and the third layer are both one-dimensional convolutional layers
with four filters of size $5$.
The second and the fourth layer are one-dimensional max-pooling layers of
size $2$.
The last layer is a fully-connected layer with $256$ units.
We apply rectified linear activation to
all of the convolutional and fully-connected layers.
Each classifier takes $d=254$ input features produced
by the GP adapter.